\documentclass[11pt]{article}

\usepackage[preprint]{acl}

\usepackage{times}
\usepackage{latexsym}

\usepackage[T1]{fontenc}

\usepackage[utf8]{inputenc}

\usepackage{microtype}

\usepackage{inconsolata}

\usepackage{graphicx}

\usepackage{enumitem}
\usepackage{paralist}
\usepackage{booktabs}
\usepackage{tabularx}

\newcolumntype{H}{>{\setbox0=\hbox\bgroup}c<{\egroup}@{}}
\usepackage{algorithm}
\usepackage{algorithmic}
\usepackage{amsmath}

\newcommand{\journal}[1]{}

\title{Beyond Top-$k$ Skill Retrieval: Diversity-Aware Skill Routing for LLM Agents}

\author{Wang Wei$^1$, Tiankai Yang$^2$, 
Samyadeep Basu$^3$, Hongjie Chen$^4$,\\
\bf{Yue Zhao$^2$, Zhengzhong Tu$^5$, Xiyang Hu$^6$, Franck Dernoncourt$^3$,} \\
\bf{Ryan A. Rossi$^3$, Hoda Eldardiry$^1$\thanks{Corresponding author.}}\\
$^1$Virginia Tech, $^2$University of Southern California, $^3$Adobe Research,\\ $^4$Dolby Labs, $^5$Texas A\&M University, $^6$Arizona State University\\
\texttt{\{wangwei718,hdardiry\}@vt.edu}
}

\begin{document}
\maketitle
\begin{abstract}
Large language model (LLM) agents increasingly rely on external skills, but routing user requests over large skill registries is difficult because many skills are functionally redundant while complex tasks often require complementary skill sets. Existing skill routers typically rank candidates independently by query relevance, which can waste context budget on redundant skills. We propose Diverse Skill Routing (DSR), a diversity-aware reranking framework that uses a Determinantal Point Process to balance relevance and non-redundancy. DSR introduces a query-residual diversity kernel that penalizes redundant skill overlap while reducing penalties caused only by shared query relevance. On the SkillRouter benchmark, DSR improves recall and full coverage over a strong pointwise reranking baseline, with larger gains on multi-skill queries. These results suggest that skill routing should be treated not only as relevance ranking, but also as complementary set selection.
\end{abstract}

\section{Introduction}

Large language model (LLM) agents increasingly rely on external tools and skills to solve tasks that require capabilities beyond direct text generation. Early tool-use systems showed that language models can learn when and how to call external APIs \citep{schick2023toolformer}, while subsequent agent frameworks use LLMs to decompose user requests, select external models or tools, and aggregate their outputs \citep{shen2023hugginggpt,qin2024toolllm}. More recently, skill-based agents organize procedural knowledge into reusable modules, such as instructions, scripts, examples, and reference documents, that can be loaded into context at inference time \citep{wang2023voyager,xu2026agentskills}. This design makes agents more extensible, since new capabilities can be added through external skill libraries rather than model retraining.

However, the growth of skill libraries creates a new routing bottleneck. When thousands or tens of thousands of skills are available, it is infeasible to expose all of them to the agent because the context window is limited, and irrelevant skills may distract execution. Recent work on skill routing studies this problem directly by retrieving task-relevant skills from large registries. SkillRouter, for example, evaluates skill selection over an approximately 80K-skill pool and shows that full skill implementations contain important routing signals beyond names and descriptions \citep{zheng2026skillrouter}. SkillsBench further highlights that curated skills can improve agent performance, but their benefit depends strongly on task and skill quality \citep{li2026skillsbench}. These findings suggest that skill selection is becoming a central component of practical LLM-agent systems.

Existing skill routers typically formulate selection as a pointwise retrieval or reranking problem: each candidate skill is scored independently against the query, and the top-ranked skills are returned. This is natural for single-skill tasks, where success depends on finding one correct skill. However, many realistic agent tasks are compositional. A user request may require several complementary skills, such as document parsing, information extraction, data transformation, and visualization. In such cases, independent top-$k$ retrieval can return redundant shortlists: several skills may have similar descriptions or implementations and therefore receive high relevance scores, while other necessary but different skills are omitted. This wastes limited context budget and weakens the agent's ability to cover all parts of a multi-step workflow.

We argue that large-scale skill routing should be treated not only as relevance ranking, but also as complementary set selection. This perspective is related to diversity-aware retrieval, where the goal is to select items that are both individually useful and mutually non-redundant. Determinantal Point Processes (DPPs) provide a principled probabilistic model for such subset selection problems by favoring sets with high item quality and high diversity \citep{kulesza2012dppfnt}. DPPs have been widely used for selecting diverse high-quality subsets, but applying them directly to skill routing is non-trivial. In skill routing, two skills may be similar because they are redundant, but they may also be similar because both are relevant to the same query while still covering different steps of the task. Penalizing all similarity uniformly can therefore remove useful complementary skills.

To address this issue, we propose Diverse Skill Routing (DSR), a diversity-aware reranking framework for large-scale skill selection. DSR builds on a standard retrieve-and-rerank pipeline: a retriever first produces candidate skills, and a quality model assigns query-dependent relevance scores. DSR then applies DPP-based selection to construct a skill set that balances relevance and non-redundancy. The key component is a query-residual diversity kernel, which measures inter-skill redundancy after removing the component of each skill representation aligned with the query. This design reduces the penalty on skills that are jointly relevant to the query, while still discouraging near-duplicate candidates. The selected skills are finally ordered by their quality scores to produce the ranked shortlist.

We evaluate DSR on the SkillRouter benchmark~\citep{zheng2026skillrouter}, which contains approximately 80K candidate skills and includes both single-skill and multi-skill queries. Compared with a strong pointwise SkillRouter baseline, DSR improves recall and full coverage, with larger gains on multi-skill queries and at larger cutoffs. Ablations show that the query-residual kernel is critical: replacing it with a standard inter-skill similarity kernel substantially reduces multi-skill full coverage. These results suggest that, as skill registries continue to grow, effective routing should account for both relevance to the user request and diversity across the selected skill set.

Our contributions are as follows:
\begin{compactitem}
    \item We formulate large-scale skill routing as a diversity-aware subset selection problem, motivated by redundancy in large skill registries and the compositional structure of multi-skill agent tasks.
    \item We propose DSR, a DPP-based reranking framework that balances query-dependent skill relevance with inter-skill non-redundancy.
    \item We introduce a query-residual diversity kernel that distinguishes redundant overlap from similarity induced by shared query relevance.
    \item We show that DSR improves recall and full coverage over a strong pointwise SkillRouter baseline, with larger gains on multi-skill queries.
\end{compactitem}

\section{Related Work}
\label{sec:related_work}

\paragraph{Tool use and skill-augmented agents.}
A growing line of work studies how LLMs can use external tools and procedural knowledge to solve tasks beyond direct text generation. Toolformer shows that language models can learn when and how to call external APIs through self-supervised training signals~\citep{schick2023toolformer}. HuggingGPT uses an LLM as a controller to decompose user requests, select expert models from Hugging Face, execute subtasks, and aggregate the results~\citep{shen2023hugginggpt}. ToolLLM scales tool learning to thousands of real-world APIs by constructing ToolBench and training models for tool-use decision making~\citep{qin2024toolllm}. Voyager studies an embodied setting where an LLM-powered agent grows an executable skill library over time and retrieves relevant skills for new tasks~\citep{wang2023voyager}. These works demonstrate the value of external tools and skills, but they do not directly address how to select non-redundant skill sets from very large skill registries.

\paragraph{Skill routing and skill evaluation.}
Recent work has begun to study skills as a first-class abstraction for LLM agents. \citet{xu2026agentskills} describe agent skills as composable packages of instructions, code, and resources that can be loaded on demand. SkillRouter directly studies large-scale skill selection, showing that routing over tens of thousands of skills is difficult and that full skill implementations provide important routing signals beyond names and descriptions~\citep{zheng2026skillrouter}. SkillsBench evaluates whether curated skills improve downstream agent performance and finds that skills can be beneficial, but their effects vary across tasks and domains~\citep{li2026skillsbench}. Our work is complementary to these studies. Rather than introducing a new skill benchmark or skill representation, we focus on the selection objective: given a candidate pool and relevance scores, how should the router construct a non-redundant shortlist for multi-skill tasks?

\paragraph{LLM routing.}
Routing across LLMs has emerged as a practical approach for improving performance under heterogeneous model capabilities and inference costs. Cost-aware systems such as FrugalGPT use cascaded routing to reduce inference cost while maintaining task performance~\citep{chen2023frugalgpt}. Other methods learn query-dependent model selection policies or model representations. EmbedLLM learns compact representations of LLMs that can support downstream applications such as model routing~\citep{embedllm}, while RouterDC uses dual contrastive learning to route queries to suitable LLMs~\citep{chen2024routerdc}. Recent work also studies routing benchmarks and learning settings, including RouterBench~\citep{hu2024routerbench}, RouterEval~\citep{routereval}, RouteLLM~\citep{ong2025routellm}, and BaRP~\citep{barp}. These methods motivate routing as a practical mechanism for efficient LLM deployment. However, LLM routing usually selects one model, a cascade, or a small set of models, whereas skill routing often needs to expose several complementary skills to an agent at once. This makes redundancy among selected items a central concern in skill routing.

\paragraph{Diversity-aware subset selection.}
Diversity has long been studied in retrieval, recommendation, and summarization. Maximal Marginal Relevance balances query relevance with novelty to reduce redundancy in reranked document lists~\citep{carbonell1998mmr}, while later diversification methods explicitly model multiple query aspects~\citep{santos2010xquad}. DPPs provide a probabilistic framework for subset selection problems that require balancing item quality and diversity~\citep{kulesza2012dppfnt}. They have been used in settings such as document summarization~\citep{cho2019multi_document}, recommendation~\citep{wilhelm2018practical_recommendation}, and information retrieval~\citep{affandi2014learning_information_retrieval,deng2020personalized_information_retrieval}. DPP MAP inference is generally challenging, and efficient greedy variants are commonly used for large-scale settings~\citep{han2017faster_map_greedy_dpp}. Our work brings this diversity-aware perspective to skill routing, where redundancy arises from overlapping procedural functionality. Unlike standard applications that penalize raw inter-item similarity, DSR computes diversity in a query-residual space tailored to query-conditioned skill selection.

\section{Method}
\label{sec:method}

We now describe how to select a compact set of skills for a query from a large skill registry. Given a user query $x$ and a skill pool $\mathcal{S}=\{s_1,\ldots,s_N\}$, the goal is to return a ranked shortlist of $k$ skills that are both relevant to the query and non-redundant with each other. Standard top-$k$ retrieval addresses only the first requirement: it ranks each skill independently by relevance and overlooks whether the selected skills cover distinct parts of the task.

Our proposed DSR addresses this limitation by combining query-dependent quality scores with diversity-aware subset selection. It first retrieves a small candidate set, assigns each candidate a quality score, and then applies DPP-based greedy MAP selection to construct a non-redundant shortlist.

We use two types of scoring models. The first is an encoder retriever, which independently embeds the query and each skill and scores a pair with cosine similarity. The encoder retriever is used for efficient candidate retrieval from the full registry. The second is a pointwise quality model, or reranker, which takes a query-skill pair as input and outputs a relevance logit. The reranker is more expressive but is applied only to the retrieved candidate set for efficiency. In our main experiments, DSR uses reranker scores as the quality signal; we later ablate this choice by replacing reranker quality with embedding-based quality in Section~\ref{sec:ablation}.

\subsection{Candidate Skill Retrieval}
\label{sec:candidate_retrieval}
Let $\mathbf{e}_x$ denote the embedding of query $x$ and $\mathbf{e}_i$ denote the embedding of skill $s_i$. All embeddings are L2-normalized. 
We define the retrieval score as $a_i(x)=\mathbf{e}_x^\top \mathbf{e}_i$ and retrieve the $M$ highest-scoring skills:
\begin{equation}
    \mathcal{C}_x
    =
    \arg\max_{\mathcal{C} \subseteq \mathcal{S}, |\mathcal{C}|=M}
    \sum_{s_i \in \mathcal{C}} a_i(x),
    \label{eq:candidate_retrieval}
\end{equation}
where $M \ll N$. DSR applies diversity-aware selection only within $\mathcal{C}_x$, which makes reranking tractable for large skill registries.
\subsection{Quality-Aware DPP Selection}
\label{sec:dpp_selection}

For each candidate skill $s_i \in \mathcal{C}_x$, DSR requires a non-negative quality score $q_i(x)$ that measures its relevance to the query. In our experiments, this score is provided by the learned pointwise reranker used in SkillRouter. Let $h_i(x)$ denote the raw reranker logit for query $x$ and skill $s_i$. We define
\begin{equation}
    q_i(x) = \sigma(h_i(x)).
\end{equation}
This transformation maps reranker logits to non-negative quality scores, which determine the item-quality terms in the DPP kernel.

DSR constructs a DPP kernel $\mathbf{L}^{(x)}$ over $\mathcal{C}_x$:
\begin{equation}
    L^{(x)}_{ij}
    =
    q_i(x) \, \phi_x(s_i,s_j) \, q_j(x),
    \label{eq:dpp_kernel}
\end{equation}
where $\phi_x(s_i,s_j)$ is a query-conditioned similarity between skills. For a subset $A \subseteq \mathcal{C}_x$, the DPP score is
\begin{equation}
    F(A;x) = \det(\mathbf{L}^{(x)}_A),
    \label{eq:dpp_objective}
\end{equation}
where $\mathbf{L}^{(x)}_A$ is the principal submatrix indexed by $A$. The determinant favors subsets whose elements have high quality scores while avoiding redundant skill representations. DSR therefore selects
\begin{equation}
    A^\star
    =
    \arg\max_{A \subseteq \mathcal{C}_x, |A|=k}
    \det(\mathbf{L}^{(x)}_A).
    \label{eq:dpp_map}
\end{equation}
When $k=1$, this reduces to ordinary relevance ranking because $\phi_x(s_i,s_i)=1$ and $L^{(x)}_{ii}=q_i(x)^2$.

\subsection{Query-Residual Diversity Kernel}
\label{sec:residual_kernel}

A standard DPP kernel can compute $\phi_x(s_i,s_j)$ directly from inter-skill cosine similarity. For skill routing, this can be too aggressive: two skills may be close in embedding space because they are redundant, but they may also be close because both are relevant to the same query. Penalizing all similarity uniformly may remove useful skills that are jointly needed for a multi-step task.

DSR instead measures diversity in a query-residual space. Let $\mathbf{e}_x$ and $\mathbf{e}_i$ be L2-normalized query and skill embeddings. We first remove the query-aligned component from each skill embedding:
\begin{equation}
    \mathbf{r}_i
    =
    \mathbf{e}_i
    -
    (\mathbf{e}_i^\top \mathbf{e}_x)\mathbf{e}_x .
    \label{eq:residual}
\end{equation}
We then blend this residual with the original skill embedding and normalize the result:
\begin{equation}
    \tilde{\mathbf{z}}_i
    =
    \lambda \mathbf{r}_i + (1-\lambda)\mathbf{e}_i,
    \qquad
    \mathbf{z}_i
    =
    \frac{\tilde{\mathbf{z}}_i}{\|\tilde{\mathbf{z}}_i\|_2}.
    \label{eq:residual_mix}
\end{equation}
where $\lambda \in [0,1]$ controls the strength of the residual projection. This mixture focuses the diversity computation on query-orthogonal variation while retaining a small amount of the original representation for stability.

The query-conditioned similarity is
\begin{equation}
    \phi_x(s_i,s_j)
    =
    \frac{1+\mathbf{z}_i^\top \mathbf{z}_j}{2}.
    \label{eq:residual_similarity}
\end{equation}
This maps cosine similarity to $[0,1]$ and ensures $\phi_x(s_i,s_i)=1$. The resulting kernel penalizes residual overlap between skills while reducing penalties caused only by shared relevance to the query.

\paragraph{Why query-residual diversity?}
Raw inter-skill similarity treats all shared embedding directions as redundancy. This assumption is too strong for skill routing because skills required by the same query often share a query-aligned component. For example, a multi-step data analysis request may require one skill for parsing a spreadsheet, another for cleaning columns, and another for generating a visualization. These skills can be close in the original embedding space because they are all relevant to the same request, but selecting them together is still useful because they cover different parts of the workflow. A standard cosine kernel can over-penalize such skills and favor candidates that are superficially different but less useful. The query-residual kernel removes the shared query direction before computing inter-skill similarity, so the diversity term focuses on overlap that remains after accounting for relevance to the same request. This matches the goal of DSR: selected skills should be jointly relevant to the query while still contributing distinct functionality.

\begin{algorithm*}[t]
\caption{DSR Inference}
\label{alg:dsr}
\begin{algorithmic}[1]
\REQUIRE Query $x$, skill pool $\mathcal{S}$, candidate size $M$, output size $k$
\STATE Encode $x$ and skills in $\mathcal{S}$ to obtain normalized embeddings $\mathbf{e}_x$ and $\{\mathbf{e}_i\}_{i=1}^{N}$
\STATE Retrieve candidate set $\mathcal{C}_x$ using Eq.~\eqref{eq:candidate_retrieval}
\STATE Compute quality scores $q_i(x)$ for each $s_i \in \mathcal{C}_x$
\FOR{each $s_i \in \mathcal{C}_x$}
    \STATE Compute residual representation:
    $\mathbf{r}_i \leftarrow \mathbf{e}_i - (\mathbf{e}_i^\top \mathbf{e}_x)\mathbf{e}_x$
    \STATE Mix residual and original embedding:
    $\tilde{\mathbf{z}}_i \leftarrow \lambda \mathbf{r}_i + (1-\lambda)\mathbf{e}_i$
    \STATE Normalize:
    $\mathbf{z}_i \leftarrow \tilde{\mathbf{z}}_i / \|\tilde{\mathbf{z}}_i\|_2$
\ENDFOR
\STATE Construct $\phi_x(s_i,s_j) \leftarrow (1+\mathbf{z}_i^\top \mathbf{z}_j)/2$
\STATE Construct DPP kernel $L^{(x)}_{ij} \leftarrow q_i(x)\phi_x(s_i,s_j)q_j(x)$
\STATE Initialize $A \leftarrow \emptyset$
\WHILE{$|A| < k$}
    \STATE Select $s^\star$ with the largest log-determinant marginal gain
    \STATE Update $A \leftarrow A \cup \{s^\star\}$
\ENDWHILE
\RETURN Selected skills in $A$, sorted by quality score
\end{algorithmic}
\end{algorithm*}

\subsection{Greedy MAP Selection and Ranking}
\label{sec:greedy_ranking}

Exact DPP MAP inference is computationally expensive, so DSR uses greedy MAP selection. Starting from the empty set, it repeatedly adds the candidate with the largest marginal gain:
\begin{equation}
    s^\star
    =
    \arg\max_{s_i \in \mathcal{C}_x \setminus A}
    \log \det(\mathbf{L}^{(x)}_{A \cup \{s_i\}})
    -
    \log \det(\mathbf{L}^{(x)}_A).
    \label{eq:greedy_gain}
\end{equation}
The process stops when $|A|=k$.

The first greedy step preserves the top prediction from the quality model. When $A=\emptyset$, the marginal gain for $s_i$ is
\begin{equation}
    \log L^{(x)}_{ii}
    =
    \log q_i(x)^2,
\end{equation}
since $\phi_x(s_i,s_i)=1$. Thus, the first selected skill is
\begin{equation}
    s_{(1)}
    =
    \arg\max_{s_i \in \mathcal{C}_x} q_i(x).
\end{equation}
Later steps condition on the selected set and favor candidates that add complementary information. The final selected skills are sorted by quality score to produce the output ranking.

DSR applies DPP selection only to the retrieved candidate set $\mathcal{C}_x$, not the full skill pool. Greedy MAP is implemented with incremental Cholesky updates, which compute log-determinant marginal gains without recomputing determinants from scratch.

Algorithm~\ref{alg:dsr} summarizes the inference procedure. DSR retrieves candidates, computes reranker quality scores, constructs the query-residual DPP kernel, and greedily selects a compact skill set before sorting the selected skills by quality.

\section{Experiments}
\label{sec:experiments}

We evaluate whether diversity-aware selection improves skill routing over large and redundant skill registries. Our experiments are designed to answer three questions:
(i) whether DSR improves coverage of required skills compared with pointwise retrieval and reranking;
(ii) whether the gains are larger for multi-skill queries; and
(iii) whether the query-residual kernel is necessary for effective diversity-aware selection.

\subsection{Experimental Setup}
\label{sec:experimental_setup}

\paragraph{Benchmark.}
We evaluate on the SkillRouter benchmark introduced by \citet{zheng2026skillrouter}, which studies skill selection over a large registry derived from the Claude Skill Registry. The benchmark contains 75 expert-verified queries over approximately 80K candidate skills. These queries span 55 domains across 8 super-categories and include both single-skill and multi-skill tasks. The single-skill subset contains 24 queries that require one target skill, while the multi-skill subset contains 51 queries that require two to five target skills. Following the benchmark protocol, we evaluate on two robustness tiers: \textsc{Easy}, with 78,361 candidate skills, and \textsc{Hard}, with 79,141 candidate skills including 780 LLM-generated distractor skills. We report averages across both tiers unless otherwise specified.

\paragraph{Baseline.}
We compare against the full SkillRouter pipeline~\citep{zheng2026skillrouter}. SkillRouter follows a retrieve-and-rerank design: SR-Emb-0.6B first retrieves candidate skills from the full registry, and SR-Rank-0.6B then reranks the retrieved candidates using the full skill text. This provides a strong pointwise reranking baseline because each candidate skill is evaluated with a learned relevance model rather than only embedding similarity. However, the final ranking is still produced independently for each skill: the score of one skill does not depend on which other skills are also selected. As a result, SkillRouter can assign high ranks to multiple overlapping skills when they are all individually relevant to the query.

\paragraph{DSR variant.}
DSR uses the same SR-Emb-0.6B retriever and SR-Rank-0.6B relevance model as SkillRouter. The only change is the final selection step: instead of returning the pointwise top-ranked skills, DSR constructs a query-residual DPP kernel over the retrieved candidates and selects a shortlist that balances relevance and non-redundancy. This design keeps candidate generation and relevance scoring fixed, allowing us to isolate the effect of diversity-aware selection. In other words, DSR does not rely on a stronger retriever or a stronger reranker; it changes how high-scoring candidates are selected together.

\paragraph{Implementation details.}
For DSR, we retrieve the top 50 candidates before applying DPP selection. The query-residual kernel uses residual mixing coefficient $\lambda=0.85$. Greedy MAP selection is implemented with incremental Cholesky updates. We evaluate ranked outputs at cutoffs $k \in \{10,20,50\}$. Since the released SkillRouter pipeline returns 20 ranked skills, its Recall@50 and Full Coverage@50 are equal to its Recall@20 and Full Coverage@20. We report these values for completeness, and focus the main comparison on shared cutoffs as well as the coverage behavior of longer DSR shortlists. All experiments were conducted on NVIDIA A100 80GB GPUs.

\begin{table*}[t]
\centering
\setlength{\tabcolsep}{12pt}
\begin{tabular}{lcccccc}
\toprule
& \multicolumn{3}{c}{Recall} 
& \multicolumn{3}{c}{Full Coverage} \\
\cmidrule(lr){2-4}
\cmidrule(lr){5-7}
Method 
& @10 & @20 & @50 
& @10 & @20 & @50 \\
\midrule
\multicolumn{7}{l}{\textit{All queries (75)}} \\
SkillRouter 
& .705 & .754 & .754
& .520 & .560 & .560 \\
DSR
& \textbf{.712} & \textbf{.768} & \textbf{.808}
& \textbf{.527} & \textbf{.573} & \textbf{.633} \\
\midrule
\multicolumn{7}{l}{\textit{Multi-skill (51)}} \\
SkillRouter 
& .659 & .704 & .704
& .424 & .458 & .458 \\
DSR
& \textbf{.668} & \textbf{.739} & \textbf{.773}
& \textbf{.432} & \textbf{.492} & \textbf{.551} \\
\bottomrule
\end{tabular}
\caption{
Full-pipeline results on the SkillRouter benchmark. Both methods use SR-Emb-0.6B retrieval and SR-Rank-0.6B relevance scores. SkillRouter applies pointwise reranking, while DSR applies query-residual DPP selection. For SkillRouter, Recall@50 and Full Coverage@50 equal the corresponding @20 values because its released output is truncated at 20. Bold indicates the best result in each column.
}
\label{tab:main_results}
\end{table*}
\paragraph{Metrics.}
We use two primary coverage metrics. \textbf{Recall@k} measures the fraction of target skills recovered in the top $k$ predictions. \textbf{Full Coverage@k} measures whether all target skills for a query are retrieved within the top $k$ positions. Full Coverage is stricter than recall and is especially important for multi-skill tasks, where missing any required skill may prevent the agent from completing the workflow. Since our focus is complementary skill-set recovery, we report MRR@k only in the appendix~\ref{app:mrr_diagnostic} as an early-precision diagnostic.

\subsection{Main Results}
\label{sec:main_results}

Table~\ref{tab:main_results} compares DSR with the full SkillRouter pipeline under the controlled setup described above. DSR improves coverage-oriented metrics across all evaluated cutoffs. On all queries, Recall@20 increases from $0.754$ to $0.768$, and Full Coverage@20 increases from $0.560$ to $0.573$. These gains are modest at the shared cutoff, but they show that DSR can recover more target skills without changing the underlying relevance model. Appendix~\ref{app:mrr_diagnostic} further shows that DSR remains close to SkillRouter on MRR, indicating that the coverage gains do not come from a large loss in early precision.

The benefit becomes clearer for longer shortlists. At cutoff 50, DSR improves overall Recall from $0.754$ to $0.808$ and Full Coverage from $0.560$ to $0.633$. This pattern is expected: pointwise reranking can place several similar skills near the top, while DSR encourages the selected shortlist to cover different parts of the task. Since the released SkillRouter output contains 20 ranked skills, its Recall@50 and Full Coverage@50 are equal to its @20 values. We therefore treat @20 as the shared-cutoff comparison and use @50 to examine whether diversity-aware selection can construct a longer, less redundant shortlist.

DSR is especially useful for multi-skill queries. On this subset, Recall@20 improves from $0.704$ to $0.739$, and Full Coverage@20 improves from $0.458$ to $0.492$. At cutoff 50, Recall improves from $0.704$ to $0.773$, while Full Coverage improves from $0.458$ to $0.551$. These results support our main hypothesis: skill routing should not only rank individually relevant skills, but also select non-redundant skill sets that better cover the requirements of complex tasks.

\subsection{Ablation Study}
\label{sec:ablation}

Table~\ref{tab:ablation} analyzes two design choices in DSR: the diversity kernel and the quality score. The diversity kernel determines how redundancy between two candidate skills is measured. The standard cosine kernel computes similarity directly from skill embeddings, while the query-residual kernel first removes the query-aligned component and then measures similarity in the residual space. The quality score determines how strongly each individual skill is favored before diversity is considered. Our main method uses reranker quality from SR-Rank-0.6B. To test whether the quality source matters, we also include an ablation that replaces reranker quality with embedding quality, where $q_i(x)$ is computed from normalized query-skill embedding similarity.

The query-residual kernel accounts for the largest ablation effect. With reranker quality, replacing the standard cosine kernel with the query-residual kernel improves multi-skill Full Coverage@10 from $0.254$ to $0.441$. This result shows that naive diversity is not sufficient for skill routing. A raw cosine kernel penalizes all inter-skill similarity, including similarity that arises because two skills are both relevant to the same query. In multi-skill tasks, this can remove useful complementary skills that share the same task context but contribute different functionality. The residual kernel avoids this failure mode by focusing the diversity penalty on overlap beyond the query direction.

The quality signal also matters. With the query-residual kernel, replacing embedding quality with reranker quality improves Recall@10 from $0.618$ to $0.711$ and multi-skill Full Coverage@10 from $0.237$ to $0.441$. This indicates that diversity-aware selection still depends on a reliable relevance signal: if the quality scores do not identify useful candidates, the DPP has less useful material to select from. At the same time, comparing SkillRouter with the residual-kernel DSR variants shows that reranker quality alone is not enough. The strongest coverage-oriented results come from combining a learned relevance model with query-residual diversity.

\begin{table}[t]
\centering
\small
\begin{tabular}{llcc}
\toprule
Kernel & Quality & R@10 & Multi-FC@10 \\
\midrule
\multicolumn{4}{l}{\textit{Pointwise baseline}} \\
-- & Reranker quality & .705 & .424 \\
\midrule
\multicolumn{4}{l}{\textit{Standard cosine kernel}} \\
Cosine & Embedding quality & .540 & .178 \\
Cosine & Reranker quality & .534 & .254 \\
\midrule
\multicolumn{4}{l}{\textit{Query-residual kernel}} \\
Residual & Embedding quality & .618 & .237 \\
Residual & Reranker quality & .711 & \textbf{.441} \\
\bottomrule
\end{tabular}
\caption{
Ablation study of DSR design choices. All DSR variants use SR-Emb-0.6B retrieval over the top 50 candidates. Embedding quality uses normalized query-skill embedding similarity as $q_i(x)$, while reranker quality uses SR-Rank-0.6B scores. R@10 is averaged over all queries; Multi-FC@10 is Full Coverage@10 on multi-skill queries.
}
\label{tab:ablation}
\end{table}

\subsection{Analysis of Reranking Strategies}
\label{sec:multi_skill_analysis}

Table~\ref{tab:multi_skill_analysis} compares different ranking and selection strategies on single-skill and multi-skill queries. The compared methods represent increasingly expressive ways to order the retrieved candidates: embedding-only Top-$k$ retrieval, zero-shot LLM reranking, the learned SkillRouter reranker, and DSR.

On single-skill queries, Top-$k$ retrieval, SkillRouter, and DSR achieve the same Recall@10 and Full Coverage@10. This suggests that when only one target skill is required, strong pointwise relevance signals are often sufficient. The main challenge appears in the multi-skill setting, where the router must recover several required skills within the same shortlist.

Compared with Top-$k$ retrieval, SkillRouter improves multi-skill Recall@10 from $0.630$ to $0.659$ and Full Coverage@10 from $0.381$ to $0.424$, showing the value of a learned reranker. DSR further improves Recall@10 to $0.668$ and Full Coverage@10 to $0.432$. These gains are smaller than those at larger cutoffs in Table~\ref{tab:main_results}, but they follow the same pattern: diversity-aware selection helps most when the task requires multiple complementary skills. The zero-shot LLM ranker performs worse than SkillRouter and DSR, suggesting that prompt-based reranking alone is not enough for fine-grained discrimination among many similar skills.

\begin{table}[t]
\centering
\small
\begin{tabular}{lcccc}
\toprule
& \multicolumn{2}{c}{Single-skill} 
& \multicolumn{2}{c}{Multi-skill} \\
\cmidrule(lr){2-3}
\cmidrule(lr){4-5}
Method & R@10 & FC@10 & R@10 & FC@10 \\
\midrule
Top-$k$ retrieval & \textbf{.875} & \textbf{.875} & .630 & .381 \\
LLM ranker & .719 & .719 & .616 & .331 \\
SkillRouter & \textbf{.875} & \textbf{.875} & .659 & .424 \\
DSR & \textbf{.875} & \textbf{.875} & \textbf{.668} & \textbf{.432} \\
\bottomrule
\end{tabular}
\caption{
Comparison of ranking and selection strategies for single-skill and multi-skill queries. Top-$k$ retrieval ranks skills by embedding similarity without reranking. The LLM ranker uses Qwen3-8B in a zero-shot reranking setting. SkillRouter uses the learned SR-Rank-0.6B pointwise reranker. DSR applies diversity-aware selection using reranker quality and the query-residual kernel.
}
\label{tab:multi_skill_analysis}
\end{table}

\paragraph{Overall takeaways.}
Across the main comparison, ablation study, and reranking-strategy analysis, the same pattern emerges. Pointwise relevance models are effective for identifying individually useful skills, especially when a query requires only one target skill. However, they do not directly optimize coverage of a required skill set. DSR improves this coverage by changing the selection objective rather than the underlying retriever or reranker. The largest gains appear for multi-skill queries and larger cutoffs, which is consistent with the role of diversity-aware selection: it mainly improves which additional skills are included after the highest-scoring candidates have already been found.

The results also clarify when diversity should be applied. The ablation shows that diversity based on raw inter-skill cosine similarity can hurt coverage, even when the same quality scores are used. This suggests that diversity is useful only when the similarity measure reflects redundancy rather than shared relevance. The query-residual kernel provides this distinction by removing the query-aligned component before measuring inter-skill overlap. As a result, DSR encourages the selected skills to remain close to the user request while reducing overlap among the selected candidates.

This distinction is important for skill routing because multi-skill queries often require several skills that are related to the same task but not interchangeable. A router that penalizes all similarity may remove useful skills simply because they share the same task context. Conversely, a router that ignores diversity may return several high-scoring but overlapping skills. DSR addresses the middle ground: it keeps the relevance signal from the learned reranker while using query-conditioned diversity to improve coverage of complementary skills.

\section{Conclusion}
\label{sec:conclusion}

We studied skill routing for LLM agents over large and redundant skill registries. Existing skill routers typically rank candidate skills independently by query relevance, which can waste context budget on redundant skills and miss complementary skills needed for multi-step tasks. To address this limitation, we proposed DSR, a diversity-aware reranking framework that uses DPP-based subset selection to balance relevance and non-redundancy. DSR introduces a query-residual diversity kernel that penalizes redundant skill overlap while reducing penalties caused by shared query relevance.

Experiments on the SkillRouter benchmark show that DSR improves recall and full coverage over a strong pointwise reranking baseline, with larger gains on multi-skill queries. Further analysis shows that the query-residual kernel is critical for effective diversity-aware selection, and that the benefits of DSR are most clear when the router must recover multiple required skills within a limited shortlist. These findings show that the selection layer remains important even when the retriever and reranker are fixed. As skill registries continue to grow, effective agent systems will need routing methods that account not only for individual relevance, but also for redundancy and complementarity among the selected skills.

\section*{Limitations}
\label{sec:limitations}

Our evaluation follows the SkillRouter benchmark, which contains 75 expert-verified queries over a large skill registry. Although the registry is large and includes both single-skill and multi-skill tasks, the number of evaluated queries is limited. Future work should evaluate diversity-aware skill routing on broader benchmarks with more domains, more complex workflows, and different types of skills.

DSR operates after candidate retrieval. If the initial retriever does not include required skills in the candidate set, the DPP selection stage cannot recover them. Our results therefore isolate the effect of diversity-aware selection given a retrieved candidate pool, but they do not remove the need for strong skill retrieval and relevance modeling. Improving candidate generation and diversity-aware selection jointly could be an important direction for future work.

We evaluate skill routing using retrieval-based metrics such as Recall and Full Coverage. These metrics measure whether the required skills are retrieved, but they do not directly measure downstream agent execution success. In practice, an agent may still fail even when all required skills are retrieved, for example because of incorrect tool use, poor planning, or conflicts among skill instructions. Future evaluations should connect skill-set coverage with end-to-end task completion.

DSR adds a DPP-based selection step after reranking. We apply this step only to the retrieved top candidates, which keeps the overhead manageable, but the cost may still matter for latency-sensitive deployments or much larger candidate sets. More efficient approximations and adaptive candidate sizes could further reduce the cost of diversity-aware selection.

\bibliography{custom}

\appendix
\clearpage

\begin{table*}[t]
\centering
\caption{
Full-pipeline results with MRR included as an early-precision diagnostic. Both methods use SR-Emb-0.6B retrieval and SR-Rank-0.6B relevance scores. The main paper focuses on Recall and Full Coverage because multi-skill routing requires recovering complete skill sets.
}
\label{tab:mrr_diagnostic}
\resizebox{0.78\linewidth}{!}{%
\begin{tabular}{l ccc ccc ccc}
\toprule
& \multicolumn{3}{c}{\textbf{MRR}}
& \multicolumn{3}{c}{\textbf{Recall}}
& \multicolumn{3}{c}{\textbf{Full Coverage}} \\
\cmidrule(lr){2-4}
\cmidrule(lr){5-7}
\cmidrule(lr){8-10}
\textbf{Method}
& @10 & @20 & @50
& @10 & @20 & @50
& @10 & @20 & @50 \\
\midrule
\multicolumn{10}{l}{\textit{All queries (75)}} \\
SkillRouter
  & \textbf{.788} & \textbf{.790} & \textbf{.790}
  & .705 & .754 & .754
  & .520 & .560 & .560 \\
DSR
  & .784 & .785 & .785
  & \textbf{.712} & \textbf{.768} & \textbf{.808}
  & \textbf{.527} & \textbf{.573} & \textbf{.633} \\
\midrule
\multicolumn{10}{l}{\textit{Multi-skill queries (51)}} \\
SkillRouter
  & \textbf{.792} & \textbf{.793} & \textbf{.793}
  & .659 & .704 & .704
  & .424 & .458 & .458 \\
DSR
  & .788 & .790 & .790
  & \textbf{.668} & \textbf{.739} & \textbf{.773}
  & \textbf{.432} & \textbf{.492} & \textbf{.551} \\
\bottomrule
\end{tabular}}
\end{table*}
\section{MRR Diagnostic}
\label{app:mrr_diagnostic}

Table~\ref{tab:mrr_diagnostic} reports MRR together with the coverage metrics from the full-pipeline comparison. MRR measures the rank of the first retrieved correct skill, so it is mainly an early-precision diagnostic. This metric is less aligned with our main goal because multi-skill routing requires recovering the full set of required skills, not only one correct skill.

DSR remains close to SkillRouter on MRR while improving Recall and Full Coverage. On all queries, DSR obtains MRR@10 of $0.784$, compared with $0.788$ for SkillRouter. On multi-skill queries, DSR obtains MRR@10 of $0.788$, compared with $0.792$ for SkillRouter. These small differences suggest that DSR improves skill-set coverage without substantially degrading the first relevant skill position.

\section{LLM Usage}
\label{app:llm_usage}

We used ChatGPT during the preparation of this manuscript for language editing and drafting support. LLMs were not used to generate experimental results or to make final scientific decisions.
\end{document}